\documentclass[letterpaper, 10 pt, conference]{ieeeconf}  

\IEEEoverridecommandlockouts                              

\usepackage{times}
\usepackage{epsfig}
\usepackage{graphicx}
\usepackage{amsmath}
\usepackage{amssymb}
\usepackage{subfigure}
\usepackage{overpic}
\usepackage[utf8]{inputenc} 
\usepackage[T1]{fontenc}    
\usepackage{hyperref}       
\usepackage{url}            
\usepackage{booktabs}       
\usepackage{amsfonts}       
\usepackage{nicefrac}       
\usepackage{microtype}      
\usepackage{multirow}

\title{\LARGE \bf
DMotion: Robotic Visuomotor Control with Unsupervised Forward Model Learned from Videos
}

\author{Haoqi Yuan$^{2,1*}$\thanks{* Equal contribution. Author ordering is determined by coin flips.} \quad Ruihai Wu$^{1*}$ \quad Andrew Zhao$^{1*}$ \quad Haipeng Zhang$^{1}$ \quad Zihan Ding$^{4}$ \quad Hao Dong$^{1,2,3\#}$\thanks{\# Corresponding author}\\
  $^1$CFCS, CS Dept., Peking University,
  $^2$AIIT, Peking University,
  $^3$Peng Cheng Lab,
  $^4$Princeton University
}

\begin{document}

\maketitle
\thispagestyle{empty}
\pagestyle{empty}

\begin{abstract}

Learning an accurate model of the environment is essential for model-based control tasks. Existing methods in robotic visuomotor control usually learn from data with heavily labelled actions, object entities or locations, which can be demanding in many cases. To cope with this limitation, we propose a method, dubbed DMotion, that trains a forward model from video data only, via disentangling the motion of controllable agent to model the transition dynamics. An object extractor and an interaction learner are trained in an end-to-end manner without supervision. The agent's motions are explicitly represented using spatial transformation matrices containing physical meanings. In the experiments, DMotion achieves superior performance on learning an accurate forward model in a Grid World environment, as well as a more realistic robot control environment in simulation. With the accurate learned forward models, we further demonstrate their usage in model predictive control as an effective approach for robotic manipulations. Code, video and more materials are available at: \url{https://hyperplane-lab.github.io/dmotion}.
\end{abstract}

\section{INTRODUCTION}

Learning the environment model is of great significance for physical scene understanding~\cite{c-swm, psd, wu2017learning, ehrhardt2018unsupervised, visual-dynamics, janner2018reasoning}, model-based reinforcement learning~\cite{watters2019cobra, veerapaneni2019entity, corneil2018efficient} and robotic manipulations~\cite{uns-physic-interact, experience-embed-visual-foresight}.
To learn a forward model of the environment transition, common approaches usually collect a large amount of labelled data (\emph{e.g.,} object identities, locations and motions) in the environment for supervised training. They learn the forward model by disentangling objects in the environment~\cite{watters2019cobra, veerapaneni2019entity}, learning compositional structures in the environment~\cite{c-swm, veerapaneni2019entity}, or learning physical properties~\cite{wu2017learning, janner2018reasoning}.

However, collecting such labelled data from the environment is costly and difficult, especially when the actor is human. It is desirable to design unsupervised methods that can learn the environment model from unlabelled observations like videos.

Existing research in cognitive science demonstrates the capabilities of infants for understanding the physical world and making predictions via unsupervised visual observation. ~\cite{internal-physic-model, infant-physical} shows that by observing moving objects in the world, infants can acquire self-awareness and build internal physics models of the world. Such physical models help humans to acquire the ability to predict the outcome of physical events and control tools to interact with the environment~\cite{core-knowledge, sensorimotor}.

\begin{figure}[h]
    \begin{center}
        \includegraphics[scale=0.35, trim={0cm, 5.5cm, 9.5cm, 0cm}, clip]{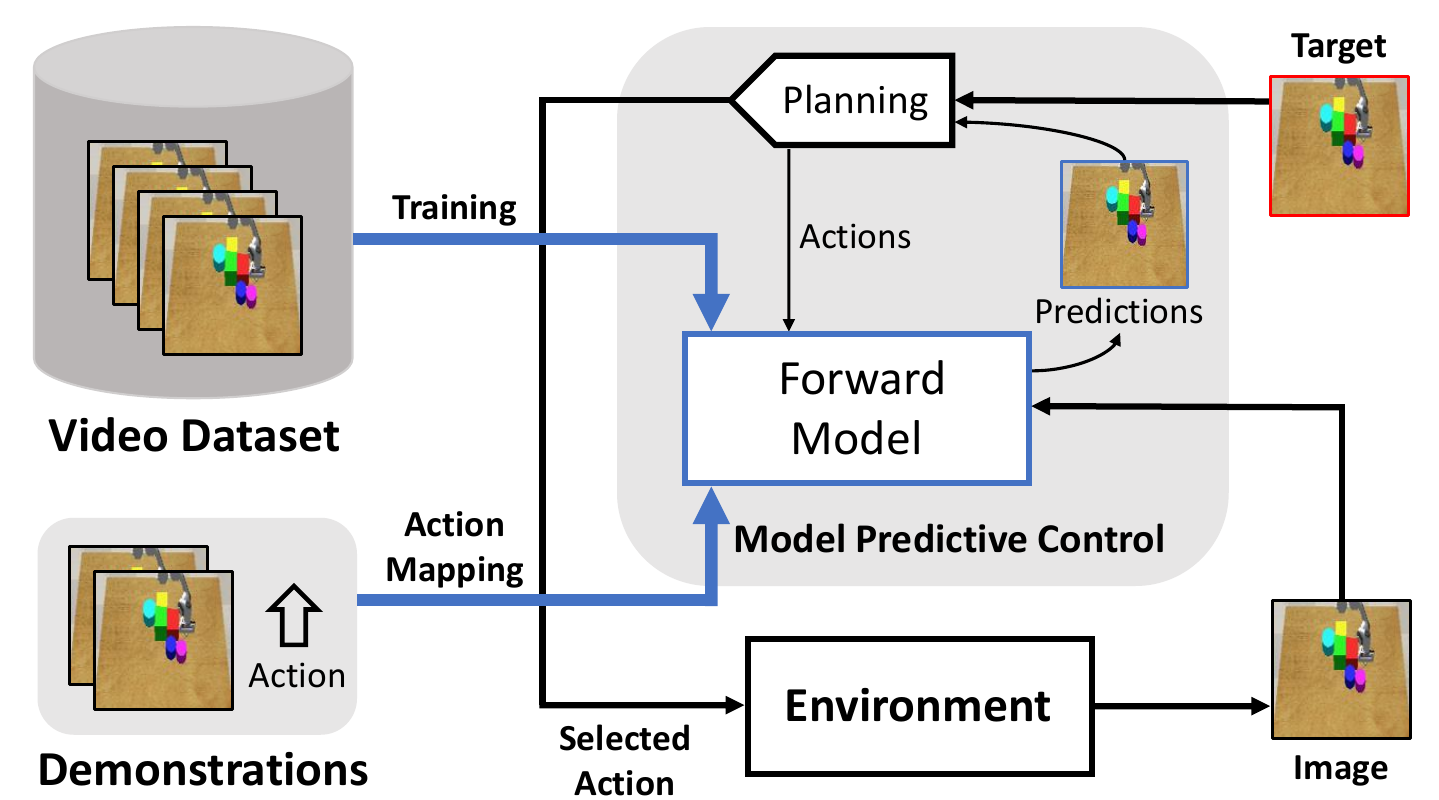}
    \end{center}
    \caption{
    A schematic overview of our framework. As indicated by the blue bold arrows in the left part, we first learn the forward model using a collected unlabelled video dataset and a few demonstrations labelled with actions. In the right part with black arrows, combined with a planning algorithm, we use the learned forward model for model predictive control (MPC). The planning algorithm generates action sequences, queries the forward model for imagined future observations, and selects the best action to reach the target in the environment.
    }
    \label{fig:motivation}
\end{figure}

Motivated by these human capabilities, in this paper, we present DMotion, an unsupervised method to learn the forward model of a given environment. Given the unlabelled videos, our model can be trained to identify the agent in an input image and predict the future observations given different agent's motions.
We observe that the essential difference between the agent and other objects is that the agent's motion is controlled by users and thus is correlated with external control signals, while other objects' motions are predictable once the agent's motion is determined. 
DMotion models the objects' motions using spatial transformers~\cite{spatial-transformer} and apply inductive bias for the motion disentanglement to identify the agent as a special object. Simultaneously, a forward model learns to predict the future observations conditioned on the disentangled agent’s motion. After unsupervised training, we only need a few video demonstrations with labelled actions to align the action space to the agent’s motion.

To verify the practicability of DMotion in real applications, we test our method on robot visuomotor control in a simulated environment. We design a model-predictive control (MPC) framework, combining our learned forward model with a planning algorithm, to perform goal-based robot pushing tasks. Figure~\ref{fig:motivation} is a schematic overview of our proposed framework.

The main contributions of our study are summarised as follows:
	1) We proposed a method that learns to disentangle the motion of the controllable agent using raw video data only. A forward prediction model of the environment is learned to predict future observations conditioned on the agent's motion, which can be directly mapped from interpretable action signals with a few demonstration samples by leveraging spatial transformation matrices.
	2) Based on the learned forward prediction model, we leverage MPC to conduct robotic visuomotor control for goal-based planning tasks in simulation. 
	3) We demonstrate in experiments that our proposed unsupervised learning (with a few demonstrations) method can achieve comparable performances as fully supervised methods, and advantageous performances over unsupervised or supervised methods with less training data, in both video prediction and visuomotor control.

\section{RELATED WORK}
Model predictive control (MPC)~\cite{mayne1999model, mayne2014model} is widely applied in robotics as a traditional approach for control, which has been testified to be effective in many robotic control tasks, including quadrotor control \cite{us2003decentralized}, dynamic legged robot control ~\cite{di2018dynamic}, robotic arm manipulations \cite{obj-centric-mpc}, SLAM of locomotory robots~\cite{tanguy2019closed}, autonomous aerial vehicles~\cite{zhang2016learning}, etc. The key insight of MPC is to continuously solve an optimisation problem online over a short horizon by making use of a system dynamic model to predict future states, including the controllable robot and its environment. Recently, people leverage machine learning techniques to bring MPC to larger-scale control problems, \emph{e.g.}, visuomotor control for robotic arms~\cite{obj-centric-mpc, ebert2017selfsupervised,7989324}. 

In visuomotor robotic control tasks, the underlying states of the robot and environment are usually not observable. The observation emission function of the system generates images from the states. Therefore, the dynamics model required by MPC is not directly accessible in general cases, which raises difficulty for applying MPC. Some works use visual affordance as auxiliary to estimate and adjust the joint configuration of robots in different tasks~\cite{reaching, grasp}. There are also works attempting to directly optimise action trajectories via raw input images~\cite{policy}, by learning embeddings without supervision~\cite{DPN,UPN, affordance}. However, the interaction among objects are not considered in these works. Some works explicitly model the dynamics of robot components and the surrounding scene, but cannot be applied to various tasks.

Forward model learning is essential for visuomotor robotic control, which can be viewed as a conditional video prediction task. Previous works have applied autoencoders with long short-term memory (LSTM) networks to achieve forward prediction in videos~\cite{lee2018stochastic, wichers2018hierarchical}. They either decompose the scene into the salient content and the motions of objects~\cite{uns-disentangle-video-pred, mcnet}, or utilise both high-level semantic information and low-level features (\emph{e.g.}, colours or edges)  for more accurate predictions~\cite{video-pred-adv-transform}. Objects disentanglement from pixels is commonly applied in video prediction tasks~\cite{wu2017learning, veerapaneni2019entity, ddpae}, even with the interactions among objects considered~\cite{wu2017learning, janner2018reasoning, veerapaneni2019entity}. Some other studies adopt action signals~\cite{watters2019cobra, veerapaneni2019entity} or other physical properties~\cite{wu2017learning, ehrhardt2018unsupervised, experience-embed-visual-foresight} in the video prediction process. However, action signals are not usually accessible in video dataset, which requires a method without action supervision. 

Several works fall in the category of unsupervised learning for video prediction model are also proposed in recent years. 
Object representations can be learned based on object segmentation results~\cite{janner2018reasoning}, by recurrently predicting attention maps of objects~\cite{burgess2019monet}, or by using a differentiable clustering method~\cite{van2018relational, greff2017neural}. Some other works disentangle objects by modelling the distributions of different objects in the scene~\cite{veerapaneni2019entity, nash2017multi, greff2019multi}.
Besides, there are some works discovering the motions~\cite{psd, visual-dynamics}, landmarks~\cite{uns-obj-landmark-cond-imggen, uns-obj-keypts-rl} or parts and structures~\cite{psd} of different objects using raw video records only.
In contrast to those methods, our model can identify the controllable agent from all objects in the environment within videos in an unsupervised manner. To achieve this goal, we utilise spatial transformer (STN)~\cite{spatial-transformer} that can effectively disentangle the motion information of objects. STN is a differentiable module that performs affine transformation matrices to feature maps conditioned on the feature map itself, allowing explicit spatial manipulation of data within the networks.  A commonly-used sampling-based optimisation method, cross-entropy method (CEM)~\cite{cem}, is applied in our method to optimise the trajectory with MPC.

\section{METHOD}

\begin{figure*}[h!]
    \begin{center}
        \includegraphics[scale=0.7, trim={1.5cm, 9.3cm, 10.5cm, 0.5cm}, clip]{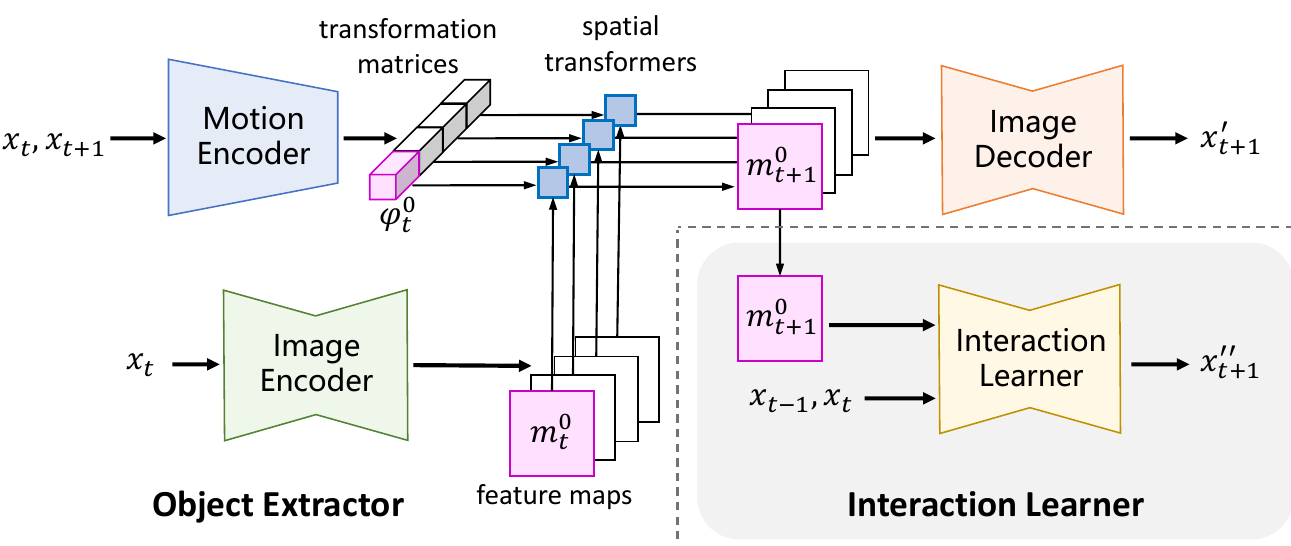}
    \end{center}
    \caption{The training process of DMotion. The model has two modules: an object extractor and an interaction learner. The object extractor consists of a motion encoder, an image encoder and an image decoder. By modelling the motion of different feature maps, the object extractor disentangles different objects in images.
    The interaction learner, indicated by the grey background, predicts the future frame conditioned on the last two frames and a feature map $m^0_{t+1}$. By training two modules together, we force the feature map $m^0$ and the transformation matrix $\varphi^0$ to contain the spatial and motion information of the agent, respectively.
    }
    \label{fig:overall_model}
\end{figure*}

We present our method for visuomotor robotic control based on a forward prediction model, which is learned from videos without labels and a few action demonstrations. We first formulate the problem of MPC for planning in Sec.~\ref{subsec:formula}. Detailed descriptions of modules for learning the forward prediction model are provided in Sec.~\ref{subsec:forward}, including an object extractor, an interaction learner and an action-transformation mapping process. With the forward model, we further demonstrate using MPC for planning as in Sec.~\ref{section-MPC}.

\subsection{Problem Formulation}
\label{subsec:formula}
The environment for robotic/object manipulation can usually be considered as a Markov decision process, which is represented as a tuple $(\mathcal{S}, \mathcal{A}, \mathcal{P})$. $\mathcal{S}$ and $\mathcal{A}$ are the state space and action space. Each state $s\in\mathcal{S}$ reflects the position and velocity information of all objects within the scene. The agent takes an action $a\in \mathcal{A}$ based on current state $s$ and change it to the next state $s^\prime$ according to a deterministic transition function $\mathcal{P}(s^\prime|s, a)$: $\mathcal{S}\times\mathcal{A}\rightarrow\mathcal{S}$.

In the settings with visual observations, the environment is usually partially observable, where the underlying state $s$ is estimated from the observation $x\in\mathcal{X}$, and $\mathcal{X}$ is a RGB image space. We aim to build a forward model $\mathcal{F}$: $\mathcal{X}\times\mathcal{A}\rightarrow\mathcal{X}$ to mirror the original transition function, such that the agent is capable of predicting the future state given the current state and action, as well as selecting actions to reach goal states using MPC. However, in the training dataset containing videos only, no additional annotations is provided. In this paper, an unsupervised learning method is proposed with spatial transformers to disentangle the motion of the agent and model the interactions between the agent and other objects, and further applied in visuomotor robotic control tasks. 

\subsection{Unsupervised Learning of Forward Models}
\label{subsec:forward}
Fig.~\ref{fig:overall_model} shows the training process of DMotion. 
The object extractor learns to disentangle the motion of the agent and other objects. Meanwhile, the interaction learner models the interaction between the agent and other objects, becoming a forward model of the environment.

\textbf{Object Extractor}
To model the interaction between the agent and other objects, where the agent is treated as a special object, we first disentangle this object from others in image observations.
Recent works~\cite{psd, visual-dynamics} show that modelling the motion of different objects separately helps object discovery and future frame prediction. %
Motivated by this, we design an object extractor module for object disentanglement.
Fig.~\ref{fig:overall_model} shows the details of the model, it
consists of an object extractor and an interaction learner.
The object extractor makes use of motion information between consecutive frames to transfer frame $x_{t}$ to the next frame $x_{t+1}$.
Specifically, the image encoder receives an image $x_t$ and outputs $n$ feature maps $\{m_t^i\}_{i=0}^{n-1}$, which are designed to represent the spatial information of all the objects in the scene. The motion encoder receives two consecutive frames $\{x_{t}, x_{t+1}\}$ to output $n$ affine transformation matrices $\{\varphi_t^i\}_{i=0}^{n-1}$, representing the spatial movement of all the objects from time $t$ to $t+1$. Each $\varphi$ is a $2\times 3$ matrix, where the first 2 columns represent rotation and scaling and the last column represents translation.
Then, each matrix transforms one corresponding feature map from the outputs of image encoder using a spatial transformer~\cite{spatial-transformer}:
\begin{equation}
    m_{t+1}^i = ST(m_t^i, \varphi_t^i), i=0,...,n-1
\end{equation}
where $ST(\cdot)$, the spatial transformation function, roughly transforms each pixel of the input feature map $m_t^i$ to its new location according to the affine transformation $\varphi_t^i$, to construct the output feature map $m_{t+1}^i$.
The image decoder then uses all the transformed feature maps $\{m_{t+1}^i\}_{i=0}^{n-1}$ to generate the next frame $x_{t+1}$.
Since the spatial transformers perform affine transformation to the feature maps, \emph{i.e.,} translation, rotation and scaling, the object extractor gives a strong inductive bias to encourage the image encoder to disentangle the spatial information of different objects into different feature maps, and the motion encoder models the motion of different objects separately.

Here, we use spatial transformers instead of cross convolutions in~\cite{psd, visual-dynamics} to transform the feature maps.
The reason is that,
for cross convolution, the range of objects' movement it can model is limited by the convolutional kernel size. Also, it is unable to model the rotation and scaling of objects explicitly.
Compared to cross convolution, spatial transformers can explicitly represent any affine transformation. Using spatial transformers, we achieve a controllable architecture to explicitly model the motion of objects. When there is no rotation and scaling in the scene, we can fix 4 parameters of the transformation matrix to enable translation only.

\textbf{Interaction Learner}
Predicting the future frame $x_{t+1}$ requires the current observations $x_{t},x_{t-1}$ of the environment and the motion of the agent at the next time step.
As shown in Fig.~\ref{fig:overall_model}, the interaction learner first takes the last two frames $\{x_{t-1}, x_{t}\}$ that contain the current state, \emph{i.e.,} position and velocity information of objects.
To provide the agent's motion at the next time step, the interaction learner takes an arbitrarily chosen feature map from the output of spatial transformers as additional input.
As is shown in Fig.~\ref{fig:overall_model},  we choose the first feature map $m_{t+1}^0$ as an input of the interaction learner.
By training the object extractor and the interaction learner jointly, we encourage the chosen transformed feature map $m_{t+1}^0$ to contain the spatial information of the agent at time step $t+1$, and the feature map $m_t^0$ contains the spatial information of the agent at time step $t$. Therefore, with the prediction loss minimised, the transformation matrix $\varphi_t^0$ should represent the motion of the agent, In this way, both the agent's location and motion are disentangled from the surrounding environment.

Overall, the object extractor and interaction learner are jointly optimised using observation triplets of $\{x_{t-1}, x_t, x_{t+1}\}$. Their objective is to minimise the prediction error of $x_{t+1}'$ and $x_{t+1}''$.
The total loss is shown as follows:
\begin{equation}
	\mathcal{L} = \|x_{t+1}'-x_{t+1}\|_2^2 + \|x_{t+1}''-x_{t+1}\|_2^2
\label{eq:loss}
\end{equation}
where the first term encourages object disentanglement, and the second term encourages the first feature map to contain the agent's spatial information.

After jointly training the modules,
the spatial information of the agent at time $t$ is contained in the first feature map $m_{t}^0$, and motion of the agent is contained in the first transformation matrix $\varphi_t^0$.

\textbf{Action-Transformation Mapping}
The aforementioned learned forward model works as follows: at each time step, the image encoder takes a frame $x_t$ to extract the feature map of agent $m_t^0$, and the user gives a transformation matrix $\phi_t^0$ to represent the agent's motion. Then, $m_t^0$ is transformed by $\phi_t^0$ and fed to the interaction learner to forecast $x_{t+1}$.
Given the initial observations $x_0, x_1$, our model can forecast a sequence of future frames conditioned on the user control recursively.

Commonly, we use an action space $\mathcal{A}$ as the interface to interact with the environment.
The user sends the action label $a\in \mathcal{A}$, rather than the transformation matrix, to move the first feature map that represents the agent. For any environments where the action reflects the agent's motion, we can easily map actions in $\mathcal{A}$ to transformation matrices, since the transformation matrix $\phi_t^0$ directly represents the position change of the agent.

For environments with finite action space,
we can obtain an instance $\{x_t, x_{t+1}\}$ showing the transition caused by $a_t\in \mathcal{A}$, then compute its corresponding transformation matrix $\phi_t^0$ using the trained motion encoder. Thus, we build an `action-transformation table' $\{a,\phi^0\}$, to interact with the environment using predefined actions as the representation of the agent's motion.

\subsection{Forward Model for Planning and Control}
\label{section-MPC}

Based on the learned forward model with motion disentangling, consisting of the image encoder and the interaction learner in Fig.~\ref{fig:overall_model}, we are able to solve the motion planning and control tasks involving the controllable agent. Specifically, a motion trajectory $a_{1:L}$ starting from an initial configuration $x_0$ to a goal configuration $x_g$ can be generated with the forward model and methods like model predictive control.

Cross-entropy method (CEM) ~\cite{cem} is applied to optimise the trajectories based on sampled motions. The overall episode has a trajectory length of $L$, which is generated with consecutive motions of the controllable agent. At each time step, we randomly sample a number of $S$ short trajectories, which are of length $ H $ (thus we have $S\times H$ motions on $S$ trajectories). With each of the sampled trajectories, the forward model is applied at the $i$-th step to continuously generate the observations $ \{x^j_{i+k}|k=1,2,...,H, j=1,2,...,S\} $ afterwards. The cost of the sampled trajectory is defined as the minimum of the distances among all generated observations $ \{x_{i+k}\} $ and the target observation $x_g$, where the location of $i$-th object of $x$ is $o^{x}_{i}$, and the distance is defined as the summation of squared $\mathit{L}2$-norm of differences between all objects' locations in generated observations and their desired locations:
\begin{equation}
    C = \min_k \sum_{i} \Vert o^{x_{i+k}}_{i}- o^{x_g}_{i} \Vert^2_2
\end{equation}
The motions sequences $a_{1:k}$ are generated from a multivariate categorical distribution since they are discrete in our settings. CEM has an optimisation iteration of $\lambda$.
The first action of the trajectory with the lowest distance to the target scene $x_g$ is selected as the action at this trajectory generation time step.

\subsection{Implementation}

{Image encoder} has 7 convolutional layers to encode the input $128\times128$ RGB image into eight $64\times64$ feature maps. 
{Motion encoder} takes two $128\times128$ RGB images as the input to produce eight transformation matrices for different feature maps, using 7 convolutional layers followed by 2 fully connected layers. Each matrix has 6 values. 
{Spatial transformers} transform each feature map by the corresponding transformation matrix following the implementation of spatial transformer networks~\cite{spatial-transformer}.
{Image decoder} concatenates the eight transformed feature maps, and then applies 5 convolutional layers to generate the output image.\\
{Interaction learner} is an encoder-decoder network using 5 convolutional layers followed by 5 deconvolutional layers. The output layer is a convolutional layer with 3 channels and tanh activation.

For each convolutional layer, we use batch normalisation with Leaky-ReLU that has a slope of 0.2.
The whole framework is jointly trained to minimise the loss function described in Equation~\ref{eq:loss} using Adam optimiser~\cite{adam} with a learning rate of 0.001. We train the model for 50 epochs with a batch size of 32.

In motion planning and control experiment, the parameters of CEM are set as follows: $L$ = 100, $S$ = 50, $H$ = 5, $\lambda=4$.

\section{EVALUATIONS}


\subsection{Datasets}

\begin{figure*}[h!]
    \begin{center}
        \includegraphics[scale=0.56, trim={1.5cm, 6.5cm, 1.5cm, 4.5cm}, clip]{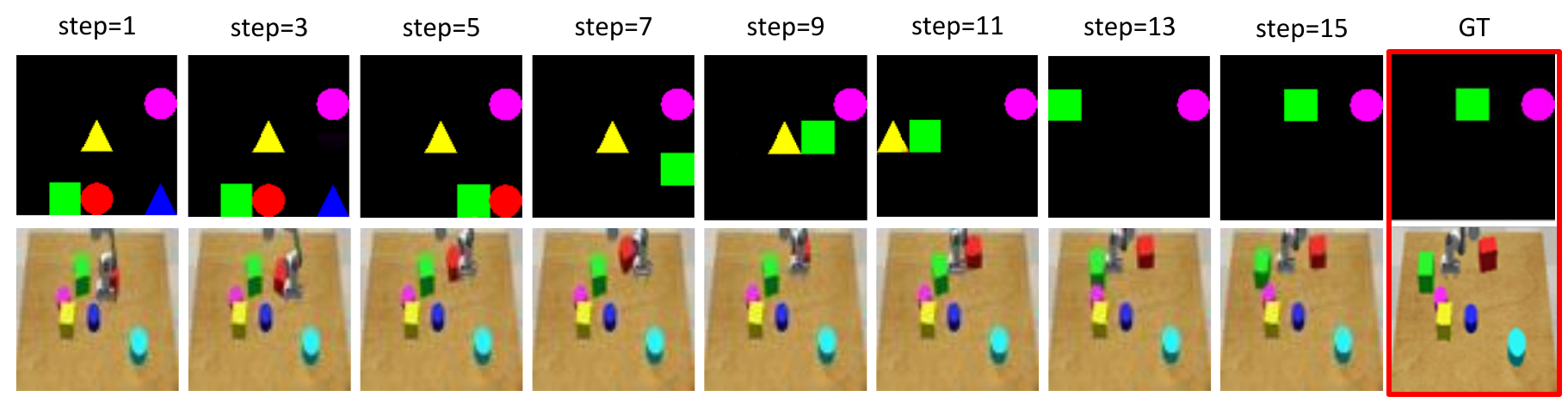}
    \end{center}
    \caption{Example results of visual forecasting conditioned on the agent's motion. First and second rows are the environments of Grid World and Robot Pushing, respectively. 
    Each row is a sample trajectory generated by our model recursively, conditioned on the agent's motion. The rightmost image in each row shows the ground truth of the last frame produced by the environment.}
    \label{fig:video}
\end{figure*}

We evaluate DMotion in two environments, as shown in 
Fig.~\ref{fig:video}.


\textbf{Grid World.} A simple environment inspired by the grid world in~\cite{c-swm}. Five different objects are placed in a $5\times5$ grid world. The green square is a controllable agent, with four possible moving directions: left, right, up, down. Other objects are passively moved due to collisions. We sample 700 trajectories, including 18,900 triplets for training and 2,100 for testing.


\textbf{Robot Pushing.} A realistic environment for robotic arm manipulations, implemented with PyRep~\cite{james2019pyrep}. 
Several different coloured objects are randomly scattered on a table. The robotic arm end-effector is also randomly initialised to be at table level. The controller can manoeuvre the robotic arm in four different directions to push any of the objects. In each trajectory, we sample either 100 timesteps or if any of the objects are pushed off the table. We created in total 6,000 trajectories or 23,8100 triplets, 214,290 triplets used for training and 23,810 used for testing. This dataset is more akin to real-world applications because the objects in the environment is subject to 3-D rotations. 

Although both Grid World and Robot Pushing can use only the current frame $x_t$ to represent the current state of the environment, we train all models with the same settings for simplicity. 
To obtain the action-transformation table, 
we randomly select one $\{x_{t}, x_{t+1}, a_{t}\}$ triplet for each of the action from the training set as demonstration.


\begin{figure*}[t!]
    \begin{center}
        \includegraphics[scale=0.8, trim={4.3cm, 5.5cm, 8cm, 5.5cm}, clip]{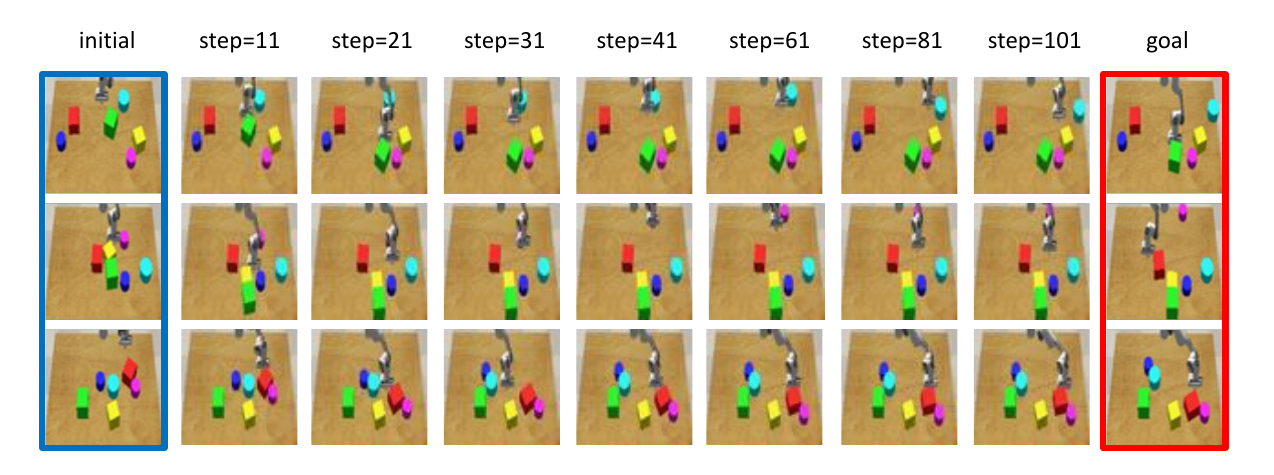}
    \end{center}
    \caption{
       Given figures of initial states (the blue boxes) and goal states (the red boxes), we visualise the states achieved by DMotion at different steps.
    }
    \label{fig:mpc_quali}
\end{figure*}


\newcommand{\tabincell}[2]{\begin{tabular}{@{}#1@{}}#2\end{tabular}}
\begin{table}[h!]
\begin{center}
\begin{tabular}{|p{1.7cm}|p{1.2cm}|p{1.2cm}|p{1.4cm}|c|}
\hline
\multicolumn{1}{|c|}{\multirow{2}{*}{\textbf{Model}}} & \multicolumn{2}{c|}{\textbf{Grid World}} & \multicolumn{2}{c|}{\textbf{Robot Pushing}}  \\ \cline{2-5} \multicolumn{1}{|c|}{}  & \textit{MSE} & \textit{Pos err.} &  \textit{MSE} & \textit{Pos err.} \\ \hline \hline


\textbf{\tabincell{c}{\footnotesize{WM AE}}} & 151\tiny{$\pm 473$} & 3.19\tiny{$\pm 17.8$} & 202\tiny{$\pm 88.81$} & 1.52\tiny{$\pm 2.94$} \\ \hline

\textbf{\tabincell{c}{\footnotesize{WM VAE}}} & 263\tiny{$\pm 642$} & 6.26\tiny{$\pm 24.3$} & 170.76\tiny{$\pm 98.52$} & 1.29\tiny{$\pm 2.7$} \\ \hline

\textbf{\tabincell{c}{\footnotesize{E-D CNN}}} & 27.4\tiny{$\pm 139$} & \textbf{0.278}\tiny{$\pm 4.85$} & 87.37\tiny{$\pm 98.63$} & 0.51\tiny{$\pm 0.81$} \\ \hline

\textbf{\tabincell{c}{\footnotesize{C-SWM}}} & 18.2\tiny{$\pm 94.9$} & \textbf{0.251}\tiny{$\pm 3.56$} & 552.18\tiny{$\pm 99.32$} & 4.46\tiny{$\pm 6.65 $} \\ \hline \hline


\textbf{\tabincell{c}{\footnotesize{WM AE}\scriptsize{(10\%)}}}   & 1169\tiny{$\pm 775$} & 19.9\tiny{$\pm 38.4$} & 276.82\tiny{$\pm 118.59$} & 2.11\tiny{$\pm 3.88$} \\ \hline

\textbf{\tabincell{c}{\footnotesize{WM VAE}\scriptsize{(10\%)}}} & 960\tiny{$\pm 724$} & 20.2\tiny{$\pm 40.1$} & 256.03\tiny{$\pm 126.79$} & 1.94\tiny{$\pm 3.82$} \\ \hline

\textbf{\tabincell{c}{\footnotesize{E-D CNN}\scriptsize{(10\%)}}} & 60.7\tiny{$\pm 230$} & 1.76\tiny{$\pm 13.8$} & 96\tiny{$\pm 99.73$} & 0.55\tiny{$\pm 0.92$} \\ \hline

\textbf{\tabincell{c}{\footnotesize{C-SWM}}\scriptsize{(10\%)}}  & 122\tiny{$\pm 338$} & 2.13\tiny{$\pm 13.1$} & 593.03\tiny{$\pm 152.99$} & 6.09\tiny{$\pm 6.65$} \\ \hline \hline


\textbf{\tabincell{c}{\footnotesize{CLASP}}} & 714.7\tiny{$\pm 355.5$} & {4.8}\tiny{$\pm 1.7$} & \textbf{81.77}\tiny{$\pm 93.94$} & 0.45\tiny{$\pm 0.78 $} \\ \hline

\textbf{\tabincell{c}{\scriptsize{DMotion-No-STN}}}  & 371\tiny{$\pm 537$} & 3.52\tiny{$\pm 15.9$} & 164.69\tiny{$\pm 136.94$} & 0.75\tiny{$\pm 1.86$} \\ \hline

\textbf{DMotion} & \textbf{14.3}\tiny{$\pm 99.8$} & \textbf{0.480}\tiny{$\pm 7.62$} & \textbf{86.78}\tiny{$\pm 132.47$} & \textbf{0.38}\tiny{$\pm 0.69$} \\ \hline
\end{tabular}
\end{center}
    \caption{Quantitative evaluation results. A lower score means better performance. 
    Baseline models masked by 10\% use 10\% of action labels in the training set. 
        }
    \label{table:eval}
\end{table}

\begin{table}[t]
\centering
\begin{tabular}{|p{1.0cm}|p{0.89cm}|p{1.16cm}|p{1.05cm}|p{1.04cm}|p{1.06cm}|}
    \hline
    \textbf{WM AE} & \textbf{\tabincell{c}{WM AE\\(10\%)}} & \textbf{E-D CNN} & \textbf{\tabincell{c}{E-D CNN\\(10\%)}} & \textbf{CLASP} & \textbf{DMotion}\\
    \hline
            0.99\tiny{$\pm 0.16$} & 1.00\tiny{$\pm 0.12$} & 0.922\tiny{$\pm 0.14$} & 1.022\tiny{$\pm 0.20$} & 0.958\tiny{$\pm 0.11$} & \textbf{0.417}\tiny{$\pm 0.16$}\\
    \hline
\end{tabular}
\caption{Quantitative evaluation results of forward models for planning and control at the final time step.}\smallskip
\label{table:quan_mpc}
\end{table}

\subsection{Baselines and Ablation Study}

\textbf{World Model (V)AE}: WM (V)AE. Inspired by ~\cite{world-model}, we first pretrain an autoencoder for the state representation using all observations. Then, a transition model learns the state transition using all action labels with the fixed autoencoder. The autoencoder can be either deterministic AE or VAE~\cite{vae}. \\
\textbf{Encoder-decoder CNN}: E-D CNN. Inspired by~\cite{uns-physic-interact}, we use an end-to-end encoder-decoder CNN architecture to predict the next frame conditioned on the last two frames and an action. \\
\textbf{C-SWM}: C-SWM ~\cite{c-swm} utilises a contrastive approach for learning structured state representations in environments. 
With the compositional structure and an additional decoder, it can learn to predict the future frames conditioned on the last two frames and the action of the agent.
\\
\textbf{CLASP}: CLASP ~\cite{clasp} uses the most similar setting to ours where training set is videos without action labels. Instead of using spatial transformers, CLASP simply uses hidden vectors to represent actions. The action representation is regularised by KL loss. \\
\textbf{Baselines with Reduced Training Set.} The above baselines, except for CLASP, are supervised learning methods, requiring samples with action labels for training, while DMotion can learn from label-free observations. 
To approximate our setting, we use a reduced training set to train the baselines, where only 10\% of action labels in the training set can be used.
Specifically, World Model (V)AE first uses all the observations to train the autoencoder, then uses 10\% of samples with action labels to train the transition model. E-D CNN and C-SWM learn from 10\% of samples with action labels. Although using only 10\% of the action labels, the baselines still use far more supervision than our DMotion.\\
\textbf{Ablation}: DMotion-No-STN. For ablation study, 
we use cross convolution operators instead of spatial transformers in the object extractor module.
We use the convolutional kernel to control the agent's motion.





\subsection{Visual Forecasting Conditioned on the Agent's Motion}

A main advantage of DMotion is to model interactions within an environment, so that it can predict the future frames based on control signals from users. To qualitatively evaluate DMotion on visual forecasting conditioned on the agent's motion, we interact with the learned environment to generate a sequence of observations.
Concretely, we initialise the environments by providing two initial frames $\{x_0, x_{1}\}$ from the test set. At each time step $t>1$, the model takes the last two frames $\{x_{t-1}, x_{t}\}$ and an action $a_t$ taken by the user to generate the next frame $x_{t+1}$. 

Fig.~\ref{fig:video} shows generated video clips for each environment. 
In Grid World, objects being pushed out are correctly predicted, and multiple objects can interact together --- \emph{e.g.,} the green square can push the red circle and blue triangle together.
The results of Robot Pushing indicates that our model could predict object rotation, multi-object interaction, and object motion when the object is partially blocked by others.
In addition, we can see that after predicting more than ten steps continuously, the generated images are still sharp and plausible.

\subsection{Quantitative Evaluation of the Forward Model}

To further demonstrate the effectiveness of DMotion, we quantitatively evaluate the accuracy of the learned forward model. 
For each model, we use all consecutive observations $\{x_{t-1}, x_t\}$ from the test set and the action $a_t$ to predict the next frame $x_{t+1}$.
We use two evaluation metrics: 
1) the mean squared error (\textit{MSE}) measures the similarity between the predicted frame and the ground truth frame, \emph{i.e.,} the error in pixel space, where the range of pixel value is scaled to $0\sim 255$.
2) the distance error of a single object is the Euclidean distance, measured by the number of pixels, between the object centres in the predicted frame and the ground truth frame, 
the position error (\textit{Pos err.}) is the averaged distance error of all objects in the frame.

The evaluation results are shown in Table~\ref{table:eval}. 
Compared to DMotion-No-STN, DMotion with spatial transformers shows significant advantages in both environments.
Even though DMotion cannot outperform all the baselines when models are trained on 100\% action supervision, 
it can still have better performance than all baselines when they are trained on the reduced training set.
Note that, even with 10\% of the action labels, the baselines still use more supervision than ours.

\begin{figure}
    \begin{center}
        \includegraphics[scale=0.45, trim={0cm, 0cm, 0cm, 0cm}, clip]{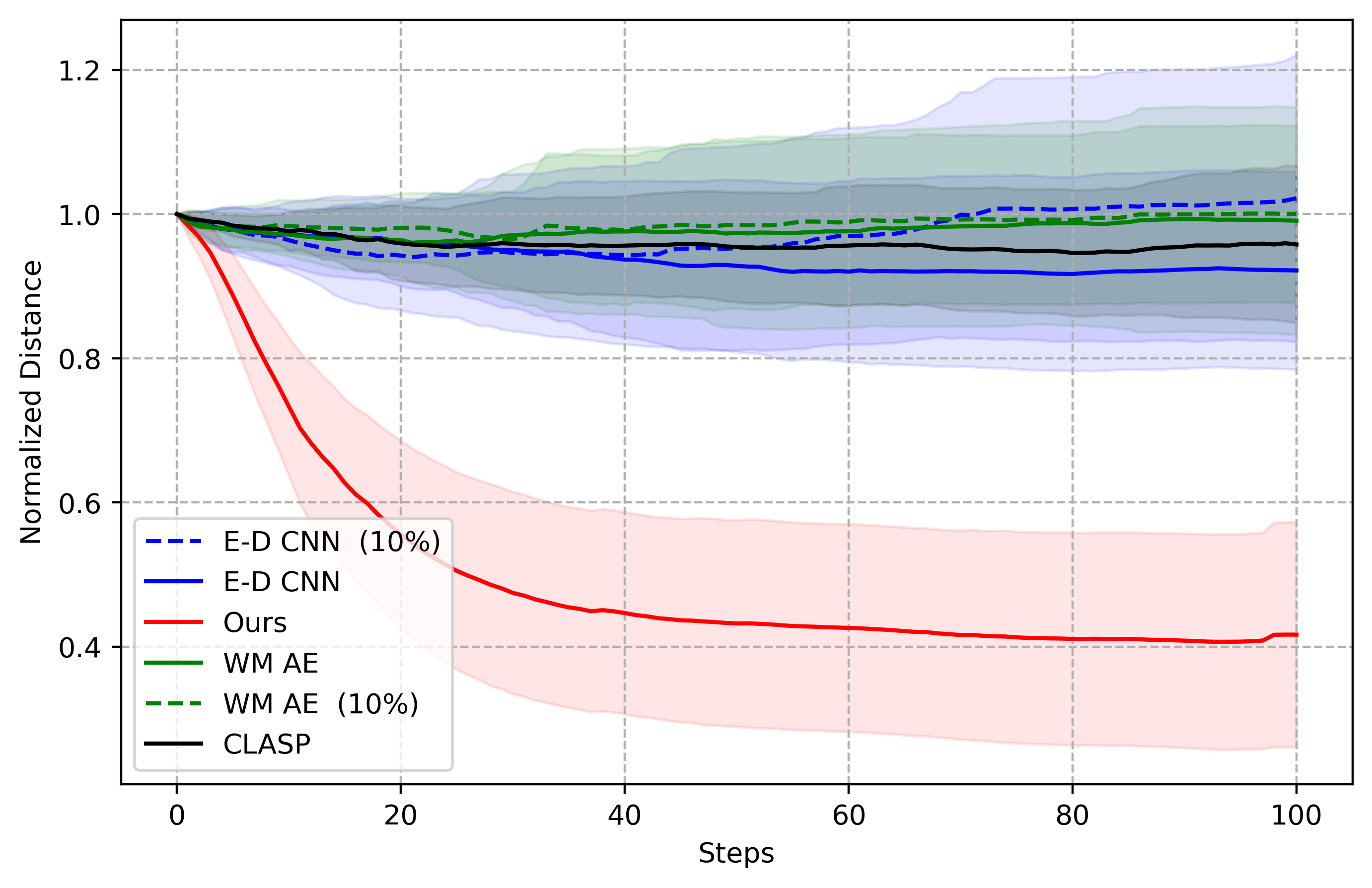}
    \end{center}
    \caption{
        Results of visuomotor control for Robot Pushing object manipulation task. The horizontal axis is the time step. The vertical axis is the average normalised distance between current and desired object locations, with the shaded regions indicating the standard deviations. Dotted lines show results from baselines trained with 10\% of labelled data.
    }
    \label{fig:mpc_quan}
\end{figure}

\begin{figure}
    \begin{center}
        \includegraphics[scale=0.38, trim={0cm, 0cm, 10cm, 0cm}, clip]{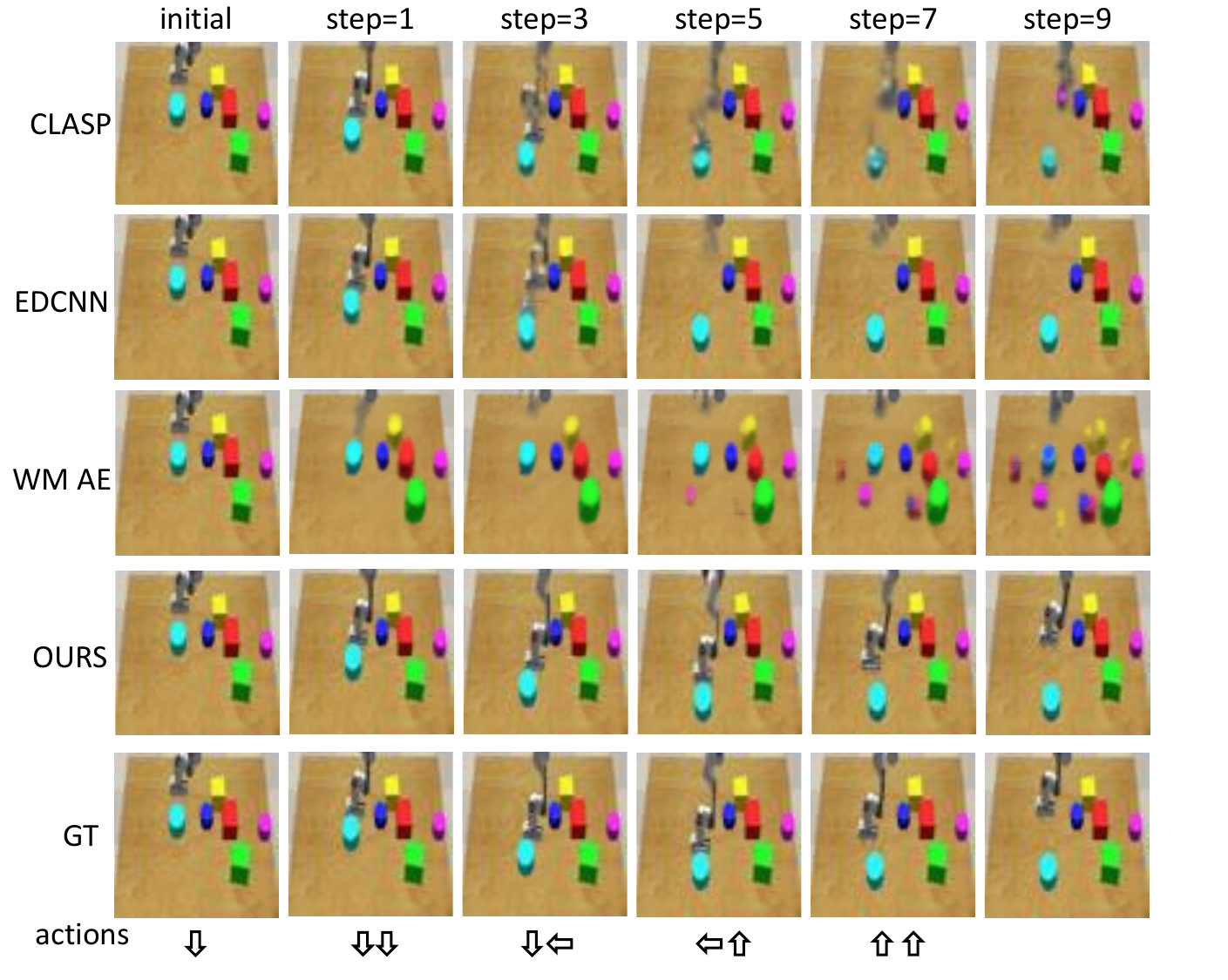}
    \end{center}
    \caption{
     Visualisation of long-term forecasting in the Robot Pushing environment. Action labels are provided at each time-step and we recursively generate the next time-step. Notice how the baselines fail to pinpoint and disentangle the exact location of the agent after a few time-steps.
    }
    \label{fig:mpc_inspection}
\end{figure}


\subsection{Evaluation of Forward Model in Planning and Control}

The task in experiments of planning and control is Robot Pushing, which contains an agent (robot arm) and 6 objects, and the goal is to propose action trajectories and control the agent to reach the target scene from the initial scene. We use our proposed forward model and CEM to optimise the action trajectories, as is mentioned in ~\ref{section-MPC}. The data is collected to guarantee that there are at least 15 interactions between the agent and other objects before deriving the target scene.

To evaluate the performance of our forward model in this task, we use the metric of the average distance of all the objects to their desired positions in the goal scene, normalised by the initial distance between the initial state and the goal state. Since the task is to achieve an arrangement of manipulable objects, the final position of the agent is not included in the evaluation for this task. Fig.~\ref{fig:mpc_quali} shows the sampled results of our forward model in this task. Fig.~\ref{fig:mpc_quan} shows the comparison between our forward model and other methods at each time step, using the above-mentioned distance metric, and Table ~\ref{table:quan_mpc} shows the quantitative results at the final time step. The forward model learned in our method with CEM is capable of achieving a much smaller average distance value (0.417) to the target positions, compared against all other methods (best 0.922). To investigate the different performances of the compared methods in planing and control, we visualize 6 of 10 frames in a consecutive visuomotor control process as in Fig.~\ref{fig:mpc_inspection}. It shows that other methods either have more and more obscure image prediction results, like WM AE, or have accurate object location prediction but obscure agent prediction, like in CLASP and EDCNN. One potential reason is that the lack of agent identification module cause the other methods to pay more attention on pixel-level reconstruction rather than agent motion prediction. DMotion based on STN achieves moderate prediction errors in recursive forecasts without losing the track of objects and the agent.


\subsection{Feature Map Visualisation}

Compared to baseline methods, DMotion is more interpretable in environment understanding, because it explicitly extracts the feature map that reflect the agent's spatial location and uses spatial transformation to describe the agent's motion. 

\begin{figure}
    \begin{center}
        \includegraphics[scale=0.5, trim={0cm, 0cm, 0cm, 0cm}, clip]{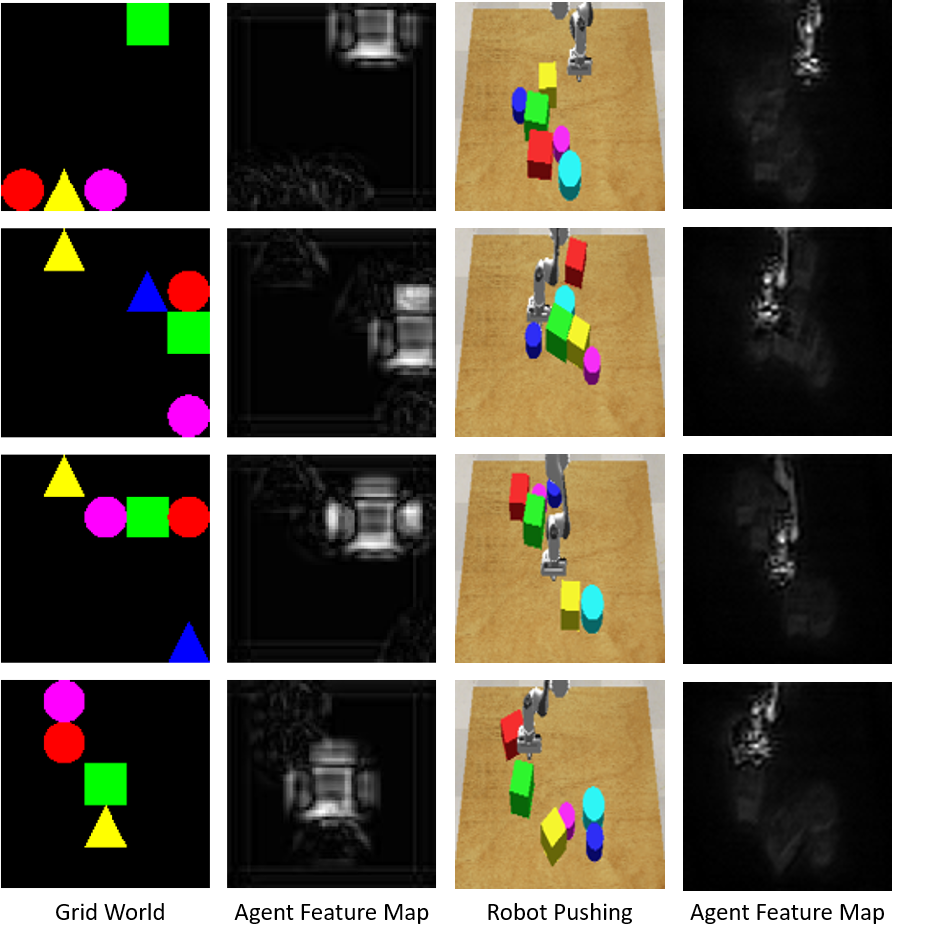}
    \end{center}
    \caption{
     Visualisation of the environment and the corresponding feature map containing the agent in the image encoder.   
    }
    \label{fig:map}
\end{figure}

To evaluate the learned agent's feature map, we visualise the first feature map from the image encoder's output.
Fig.~\ref{fig:map} shows some visualisation results of the feature maps, where the input images are randomly picked from the test set.
To have a better view, as the values of feature maps are around zero, we visualise the absolute values of the maps.
The result shows that  DMotion successfully identifies the agent in the environment, the feature maps contain the spatial information of the agent.


\section{CONCLUSIONS}
In this paper, we present DMotion for visuomotor robotic control in simulation via unsupervised learning a forward prediction model from videos, which demonstrates superior performances over other supervised or unsupervised methods in the model predictive control experiments. For video forward prediction, the proposed methods leveraging a few demonstration samples can achieve comparable performances as supervised learning methods like World Model with Auto-Encoder and Encoder-Decoder CNN, and even beat them when reducing the amount of labelled training data. As an unsupervised method, CLASP cannot perform as good as our method in visuomotor control tasks although it also has high accuracy in video prediction. Further extension of our work to 3D object manipulations and deployment on real-world visuomotor robotic control are potential future works.

\section{FUTURE DISCUSSION}
This study is the first to learn the unsupervised forward model via explicitly disentangling the agent's motion.
Motivated by human capabilities, in the future, we believe machines can learn the  model of complex environments from unlabelled observation.
To enable this framework to behave in more complex scenarios, in the future, we will need to cope with the problems including continuous action space, multi-agent, articulated agent/object and the change of physical parameters.

\section*{ACKNOWLEDGEMENT}
This project was supported by 
National Natural Science Foundation of China —Youth Science Fund (No.62006006): Learning Visual Prediction of Interactive Physical Scenes using Unlabelled Videos. 
We would also like to thanks the funding from
Key-Area Research and Development Program of Guangdong Province (No.2019B121204008)
and
National Key R\&D Program of China: New Generation Artificial Intelligence Open Source Community and Evaluation (No.2020AAA0103500), Topic: New Generation Artificial Intelligence Open Source Community Software and Hardware Infrastructure Support Platform (No.2020AAA0103501).

\bibliographystyle{IEEEtran}
\bibliography{IEEEabrv,mybibfile}

\end{document}